\documentclass[pmlr]{jmlr}% new name PMLR (Proceedings of Machine Learning Research)

\usepackage{longtable}% for long tables

\usepackage{booktabs}
 \usepackage{siunitx}

\usepackage{graphicx}
\usepackage{float}
\usepackage{multicol}
\usepackage{multirow}
\usepackage{makecell}

\theorembodyfont{\upshape}
\theoremheaderfont{\scshape}
\theorempostheader{:}
\theoremsep{\newline}

\jmlrvolume{220}
\jmlryear{2023}
\jmlrworkshop{NeurIPS 2022 Competition Track}
\editors{Marco Ciccone, Gustavo Stolovitzky, Jacob Albrecht}
\title[The NeurIPS 2022 Neural MMO Challenge]{The NeurIPS 2022 Neural MMO Challenge: A Massively Multiagent Competition with Specialization and Trade}

  \author{
   \Name{Enhong Liu\nametag{\thanks{Equal Contribution}}}  \Email{mudouliu@chaocanshu.ai} \\
   \addr Parametrix.AI \\ 
   \Name{Joseph Suarez$^*$} \Email{jsuarez@mit.edu} \\ \addr Massachusetts Institute of Technology \AND
   \Name{Chenhui You} \Email{kirstyyou@chaocanshu.ai} \\
   \Name{Bo Wu} \Email{bowu@chaocanshu.ai} \\
   \Name{Bingcheng Chen} \Email{arbingchen@chaocanshu.ai} \\
   \Name{Jun Hu} \Email{junhu@chaocanshu.ai} \\
    \Name{Jiaxin Chen\nametag{\thanks{Corresponding Author}}} \Email{jiaxinchen@chaocanshu.ai} \\
   \Name{Xiaolong Zhu} \Email{xiaolongzhu@chaocanshu.ai} \\ \addr Parametrix.AI \\ \AND
   \Name{Clare Zhu} \Email{clarezhu@cs.stanford.edu} \\ \addr Stanford University \\
   \Name{Julian Togelius} \Email{julian.togelius@nyu.edu} \\ \addr New York University \\
   \Name {Sharada Mohanty} \Email{mohanty@aicrowd.com} \\ \addr AICrowd \AND   
   \Name{Weijun Hong} \Email{ hongweijun@corp.netease.com} \\ \addr NetEase Games AI Lab \\
   \Name{Rui Du} \Email{durui@bilibili.com} \\ \addr Bilibili \\
   \Name{Yibing Zhang} \Email{554011619@qq.com} \\ \addr Chengdu Goldwin Electronics Technology Co., Ltd \\ \AND 
    \Name{Qinwen Wang} \Email{wangqinwen@bupt.edu.cn} \\
     \Name{Xinhang Li} \Email{lixinhang@bupt.edu.cn} \\ 
     \Name{Zheng Yuan} \Email{ yuanzheng@bupt.edu.cn } \\ 
     \Name{Xiang Li} \Email{ lixianglgc@bupt.edu.cn } \\ 
     \Name{Yuejia Huang} \Email{huangyuejia@bupt.edu.cn} \\\addr Advanced Network Technology Laboratory, Beijing University of Posts and Telecommunications \\ 
     \AND
     \Name{Kun Zhang} \Email{mori42@qq.com} \\\addr Independent Researcher \\ \AND
     \Name{Hanhui Yang} \Email{hhyang@math.cuhk.edu.hk} \\ \addr The Hong Kong Chinese University \\
     \Name{Shiqi Tang} \Email{sqtang2-c@my.cityu.edu.hk} \\ \addr The Hong Kong City University \\\AND
     \Name{Phillip Isola} \Email{phillipi@mit.edu} \\ \addr Massachusetts Institute of Technology
   }

\begin{document}

\maketitle

\begin{abstract}

In this paper, we present the results of the NeurIPS-2022 Neural MMO Challenge, which attracted 500 participants and received over 1,600 submissions. Like the previous IJCAI-2022 Neural MMO Challenge, it involved agents from 16 populations surviving in procedurally generated worlds by collecting resources and defeating opponents. This year's competition runs on the latest v1.6 Neural MMO, which introduces new equipment, combat, trading, and a better scoring system. These elements combine to pose additional robustness and generalization challenges not present in previous competitions. This paper summarizes the design and results of the challenge, explores the potential of this environment as a benchmark for learning methods, and presents some practical reinforcement learning training approaches for complex tasks with sparse rewards. Additionally, we have open-sourced\footnote{https://github.com/NeuralMMO/NeurIPS2022NMMO-Submission-Pool} our baselines, including environment wrappers, benchmarks, and visualization tools for future research.

\end{abstract}

\section{Introduction}
\label{sec:intro}

The ecosystems and populations of Earth feature mixed cooperation and competition. Inspired by this, the Neural MMO environment was designed to accommodate a large number of agent populations with limited resources, necessitating socially-aware planning for survival. We propose the Neural MMO Challenge as a means to explore the emergence of many-agent intelligence and promote research on cognitively sophisticated reinforcement learning environments.

We have held two Neural MMO challenges to date. The previous challenge was designed to study multi-task RL and provide a benchmark for robustness and generalization in multi-agent systems \cite{ref1}. Although it attracted many players and yielded valuable policies, the exact task definition and unbalanced attack mechanisms limited the emergence of interesting behaviors. Therefore, we require more complex environments and a sparse scoring system to make the environment more engaging.

In the NeurIPS-2022 Neural MMO Challenge, we updated the game mechanics and scoring method of the Neural MMO competition. The specific adjustments are as follows:

\begin{enumerate}
    \item In order to encourage the emergence of more interesting strategies, we do not set an explicit task for agents. Instead, we demand that agents complete a final, sparse goal: survival for as long as possible.
    \item We use the updated v1.6 release of Neural MMO, which introduces combat, professions, equipment, a battle-royale inspired fog mechanic, and a trading mechanism.
    \item We have made dozes of fixes to the environment based on the previous competition to ensure that it is fair and strategically interesting.
\end{enumerate}

Other multiagent reinforcement learning (MARL) environments include StarCraft 2 \cite{ref2} (open-source) and Dota2 \cite{ref3} (not publically available), as well as a number of smaller and more niche projects. The main issue is that no other environment couples large population capacity with high per-agent complexity. The competition build of Neural MMO features 128 agents spread across 16 teams where each individual agent must make complex decisions about foraging, combat, specialization, cooperation, and trade. The mechanics introduced by the latest version of the environment support more diversified and complicated policies compared to previous versions. Our contributions based on the NeurIPS-2022 Neural MMO competition are as follows:

\begin{enumerate}
\item We provide a specific, controlled setting of Neural MMO suitable for benchmarking a wide range of learning methods.
\item We establish an imitation learning (IL) track and associated dataset of agent trajectories. Most competitions in this area are limited to reinforcement learning and rule-based (scripted) approaches. This competition shows that IL can also achieve satisfactory results.
\item We summarize our experience in designing the competition tracks and resources for participants. We have previously held a similar competition at IJCAI, but much of our previous experience needs updating. We believe the structures we present in this paper will be useful for other competitions.
\end{enumerate}

\section{Related Work}

The real world is massively multiagent, and research on MARL is essential for solving large-scale and cognitively realistic problems. To aid in the study of multi-agent behavior, Jiang et al. proposed a basic MARL environment library with several simple grid-based environments \cite{ref4}. The particle world project \cite{ref5} allows users to customize MARL environments using their APIs. However, the rules and action space in these environments are limited.

To address this, researchers have created more realistic environments such as Fever Basketball \cite{ref6}, Google Research Football \cite{ref7}, Pommerman \cite{ref8}, and StarCraft II. The latter allows observation of the control ability of a MARL algorithm for multi-agent units and has led to the development of well-known algorithms like Qmix \cite{qmix} and COMA \cite{coma}. While these environments enable RL algorithms to control multiple agents with different action spaces, the number of agents is still far less than the number in natural populations.

Neural MMO, an open-source complex MARL environment, is unique in that it can accommodate a vast number of populations while providing limited resources for survival, forcing populations to learn to cooperate and compete. With the introduction of a battle-royale inspired fog mechanic, equipment mechanism, and trading mechanism, researchers can simulate social behaviors, evaluate the robustness and generalization of RL algorithms, and explore diverse multi-agent research topics.

\begin{figure}[!b]
    \centering
    \includegraphics[width=4.5in]{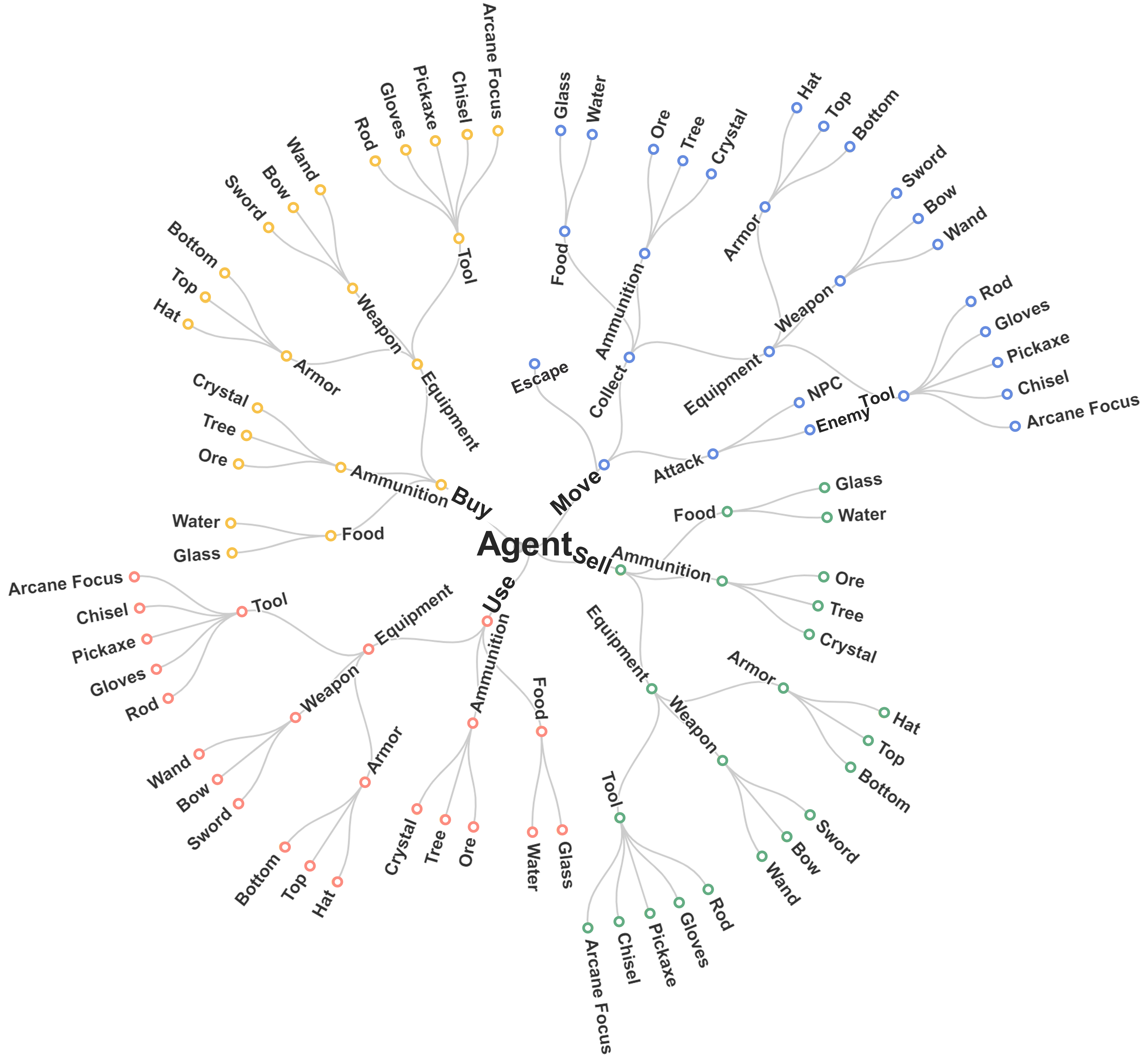}
    \caption{The decisions that agents face at every game tick of the environment }
    \label{decision tree}
\end{figure}

\section{Competition Overview}

Neural MMO is an environment where agents must fight, forage, barter, and more in order to survive against multiple teams of intelligent adversaries. As shown in Fig. \ref{decision tree}, agents must make complex decisions at each game tick. The large decision space and long-term dependency of behaviors bring diversity to agents' strategies.

The environment has the following characteristics:

\begin{enumerate}
\item \textbf{Massively multiagent:} The Neural MMO environment can accommodate 1000+ agents, a capability that few environments possess.
\item \textbf{Interesting mechanics:} The battle-royale inspired fog mechanic, combat system, equipment system, and trading mechanism are introduced in Neural MMO. These fresh game mechanics make the environment appealing to researchers and worth exploring.
\item \textbf{Flexible settings:} The generation ratio of various terrains in the environment can be adjusted at any time, which enables better testing of the stability and robustness of RL algorithms. Additionally, game mechanics can be controlled flexibly to accommodate different research topics.
\item \textbf{Sparse scoring method:} The environment uses a sparse scoring rule for ranking. In other words, agents must complete long-term tasks with unclear rewards, which encourages the emergence of interesting behaviors and diverse strategies.
\item \textbf{Support for various algorithms:} The environment supports arbitrary agent controllers, not just RL algorithms.
\end{enumerate}

At each step of the game, a map with dimensions of 128 $\times$ 128 is randomly generated, with 128 agents from 16 teams spawning around the edges of the map. Each agent begins with initialized health, food, and water levels set to 100. Teams must navigate the environment, gathering resources, equipping weapons, hunt NPCs, and engage in combat. The game includes the following features:

\begin{enumerate}
    \item \textbf{Terrain}: There are 16 different types of terrain on the map, each offering different benefits to agents who pass through them. For example, passing through foliage replenishes food, while passing through ore generates melee ammo.
    \item \textbf{NPCs}: Three types of NPCs are evenly distributed from the edge to the center of the map. NPCs play a critical role in the game, as agents primarily obtain gold and items by defeating them. Generally, the higher-level NPCs an agent defeats, the better the items they can obtain.
    \item \textbf{Blue Circle}: From the 240th game tick, a battle-royale inspired fog or "blue circle" begins to shrink. For every 16 ticks, the circle shrinks one tile. Agents within the circle take damage proportional to their distance from the safe zone.
    \item \textbf{Skills}: Agents can train eight skills. A higher skill level allows agents to use better equipment, which in turn allows them to inflict more damage on enemies, take less damage, or restore more HP/Food\&Water.
    \item \textbf{Combat}: Agents have access to melee, range, and mage attacks, which follow a rock-paper-scissors dominance relationship: melee beats range, range beats mage, and mage beats melee. Dominance is calculated using the attacker's chosen attack skill and the defender's main combat skill. The dominant style causes 1.5 times damage.
\end{enumerate}

\subsection{Observation Space}
To make decisions, each agent can observe various aspects of the game, including its own equipment information, the global information store, some enemy information, all teammate information, and all terrain information within its field of view.

\subsection{Action Space}
Table 1 shows the available actions that the agent can take. These include deciding where to move, who to attack, how to attack, what to use, what to buy, and what interaction information to expose. Note that the agent cannot move to a tile with other agents or certain special terrains, which are outlined in Table 2. Using an item will have different effects based on its properties, which are detailed in Table 3. When selling equipment, the agent can select any item in its backpack, specify the selling price, and place the item on the global market. If the sale is successful, the agent will receive the proceeds from the sale.

\begin{table}[h]
 \label{action1}
 \caption{Introduction of the action space of an agent in the Neural MMO.}
 \centering
\setlength{\tabcolsep}{9mm}{
\begin{tabular}{lll}
\toprule
\textbf{Action} & \textbf{Option}           & \textbf{Dimension} \\
\midrule
Move            & Up,Down,Left,Right        & 3           \\
Attack Target   & Enemy in sight            & 225         \\
Attack Style    & Melee,Ranged,Mage         & 3           \\
Use             & Item in package           & 12          \\
Sell            & Item in package           & 12          \\
Buy             & Item in global market     & 170         \\
Communicate     & Discrete numerical values & 128        \\
\bottomrule
\end{tabular}%
}
\end{table}

\begin{table}[t]
 \label{action2}
 \caption{Different terrains and their corresponding property.}
\centering
\setlength{\tabcolsep}{4.5mm}{
\begin{tabular}{llll}
\toprule
\multicolumn{1}{c}{\textbf{Element}} & \multicolumn{1}{c}{\textbf{Usage}} & \multicolumn{1}{c}{\textbf{Passable}} \\ 
\midrule 
Lava     & Lethal upon contact                             & False \\
                    Water    & Harvest Water                                      & False \\
                    Grass    & Can walk on                                               & True  \\
                   Scrub    & Forest degrades to Scrub after being harvested     & True  \\
                 Forest   & Harvest Forest for producing food                  & True  \\
                   Stone    & Impassible barriers                                           & False \\
                   Slag     & Ore degrades to Slag after being harvested         & True  \\
                   Ore      & Harvest ore to produce melee ammunition            & True  \\
                   Stump    & Tree degrades to Stump after being harvested       & True  \\
                   Tree     & Harvest Tree to produce range ammunition           & True  \\
                  Fragment & Crystal degrades to Fragment after being harvested & True  \\
                  Crystal  & Harvest crystal to produce mage ammunition         & True  \\            Weeds    & Herb degrades to weeds after being harvested       & True  \\
                 Herb     & Harvest Herb to produce Ration                     & True  \\
                 Ocean    & Fish degrades to ocean after being harvested   & False \\
                 Fish     & Harvest Fish to produce Poultice                   & False \\
\bottomrule
\end{tabular}}
\end{table}

\begin{table}[t]
\label{action3}
\caption{Different items and their corresponding property.}
\centering
\setlength{\tabcolsep}{3mm}{
\begin{tabular}{llll}
\toprule
\multicolumn{1}{c}{\textbf{Category}} & \multicolumn{1}{c}{\textbf{Usage}}                       & \multicolumn{1}{c}{\textbf{Items}}     & \multicolumn{1}{c}{\textbf{Level}} \\
\midrule
Ammunition & Increase damage                  & Shaving/Scrap/Shard & 1-10  \\
Weapon     & Increase damage                  & Sword/Bow/Wand      & 1-10  \\
Armor      & Increase defense                 & Hat/Top/Bottom      & 1-10  \\
Consumable & Restore Food/Water/HP            & Ration/Poultice     & 1-10 \\
Tool                              & \makecell[l]{Increase the level of \\ products after gathering resources}   & \makecell[l]{Rod/Gloves/Pickaxe \\ Chisel/Arcane Focus} & 1-10                               \\
Gold       & Buy items from the global market &                     & N/A \\
\bottomrule
\end{tabular}%
}
\end{table}

\subsection{Technical Details}
\begin{figure}[!t]
    \centering
    \includegraphics[width=5.5in]{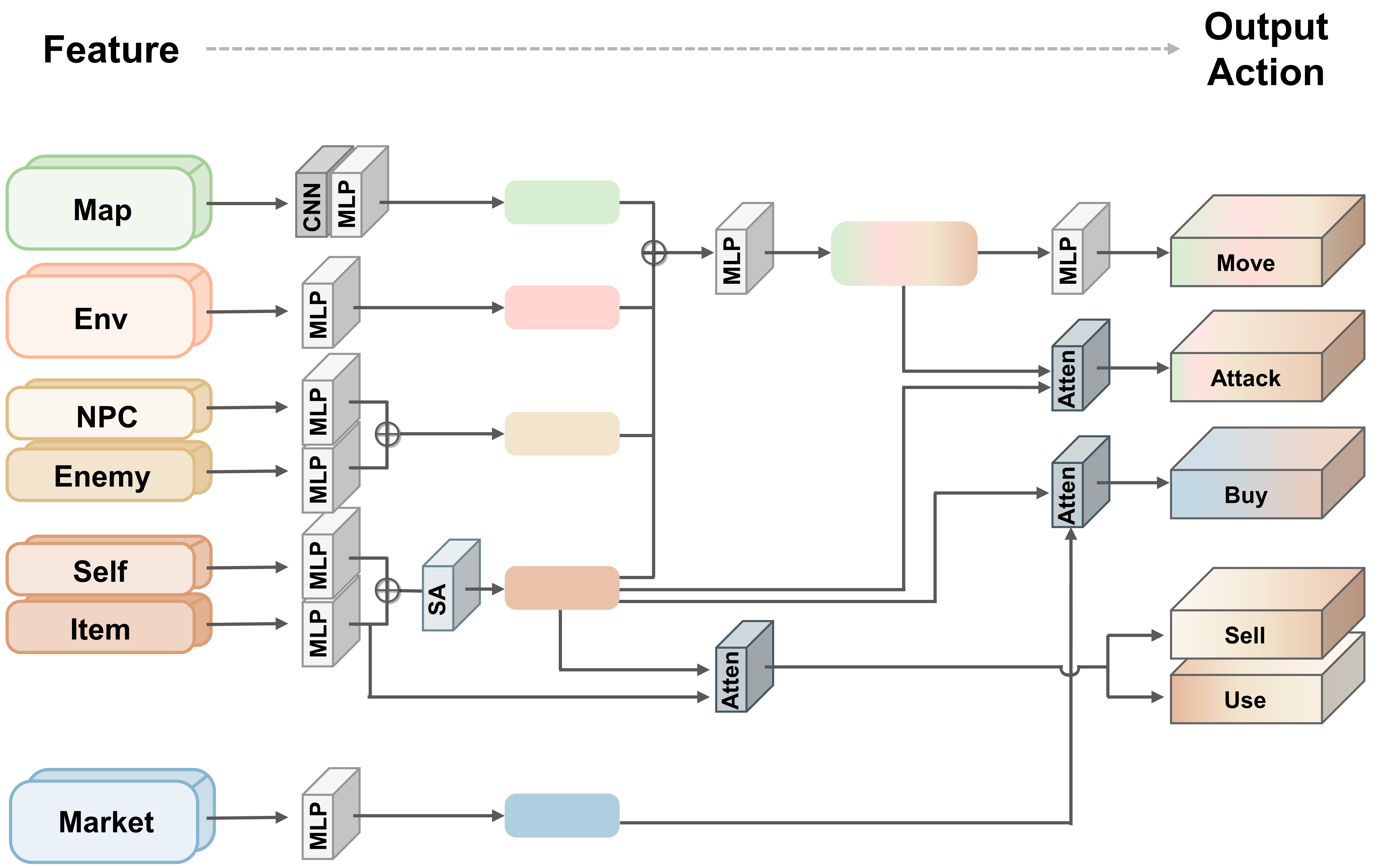}
    \caption{Model architecture for the PvE stage 2 built-in AI. Subnetworks include MLP, CNN, self-attention (SA), and attention. Baseline included with code release.}
    \label{struc}
\end{figure}

In PvE stage 2, participants face RL-based agents that we trained. Figure \ref{struc} shows the structure of the model we designed. We adopt a team-based modeling approach, which integrates the information of the entire team to obtain a joint policy. For table-type features such as self-features, enemy features, env features, and market features, we use an MLP for encoding. We apply a CNN to process the image-type features, i.e., map features. To improve data efficiency, we introduce an attention component into the model. For instance, in the buy action, buying items is related to the agent and the items existing in the market. Therefore, we perform attention to self-embedding and market embedding and put the attention embedding into the MLP to obtain the final action.

We take a curriculum learning approach to train the model and obtain three different styles of built-in AI, which we refer to as "reckless", "ruthless", and "coward". In the first training stage, we set general rewards for agents to help them learn basic skills such as attacking, gathering, and escaping the blue circle. This produces the "reckless" baseline. These agents do not cooperate well, so we continue training with an additional reward to produce the "ruthless" baseline. In addition, we also set some penalties based on the checkpoint of the first training stage and obtain the "coward" baseline. Compared to the previous two built-in AIs, the "coward" watches and reacts to situations and never strikes more capable rivals first.

\section{Competition Summary}

\subsubsection{Improvement}
\begin{figure}[!t]
    \centering
    \includegraphics[width=6in]{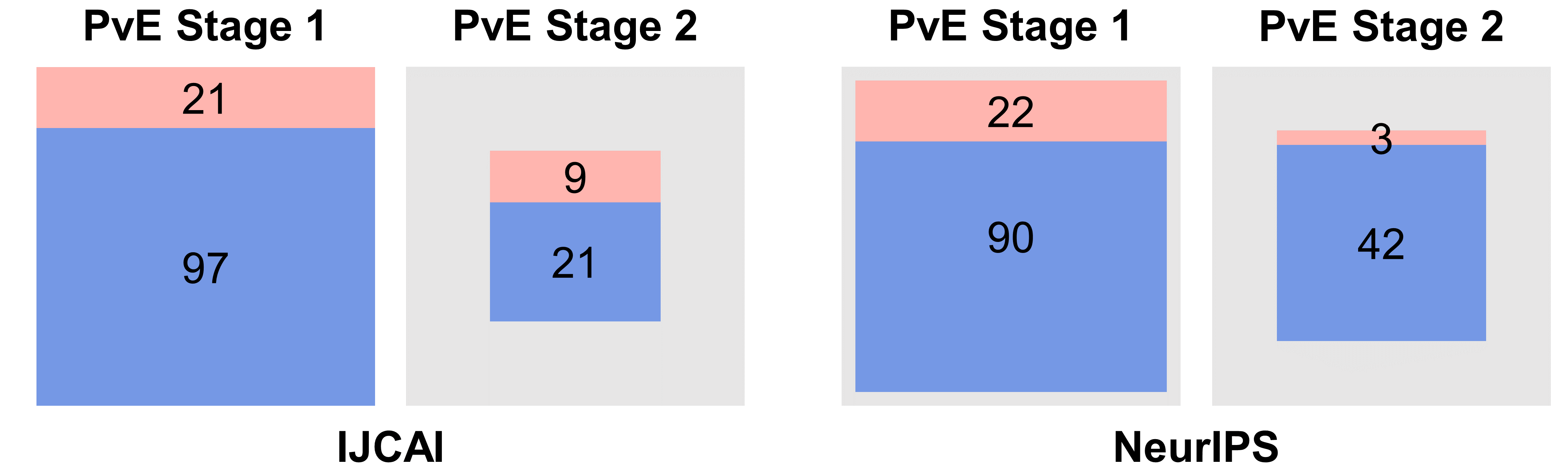}
    \caption{Performance of participants in the IJCAI and NeurIPS Neural MMO challenges. In the IJCAI competitions, 21 participants achieved 1.0 top 1 ratio in PvE stage 1 and 9 participants achieved 1.0 top 1 ratio in PvE stage 2. In the NeurIPS Neural MMO challenge, 22 participants achieved 1.0 top 1 ratio in PvE stage 1 and only 3 participants achieved 1.0 top 1 ration in PvE stage 2}
    \label{improve}
\end{figure}

The NeurIPS Neural MMO competition attracted over 20,000 views, 475 registrations, 134 team registrations, and 1651 submissions. Compared to the previous competition, this year's competition was more challenging due to the introduction of more complex game mechanics. As depicted in Fig. \ref{improve}, only a few participants achieved ideal results. In the PvE stage 1 of the IJCAI and NeurIPS challenges, many players were able to successfully overcome the built-in AIs. However, in the PvE stage 2, only 9 participants outperformed the built-in AIs in the IJCAI challenge, and in this challenge, only the top 3 participants were able to outmatch the built-in AIs.

\begin{figure}[!ht]
    \centering
    \includegraphics[width=6in]{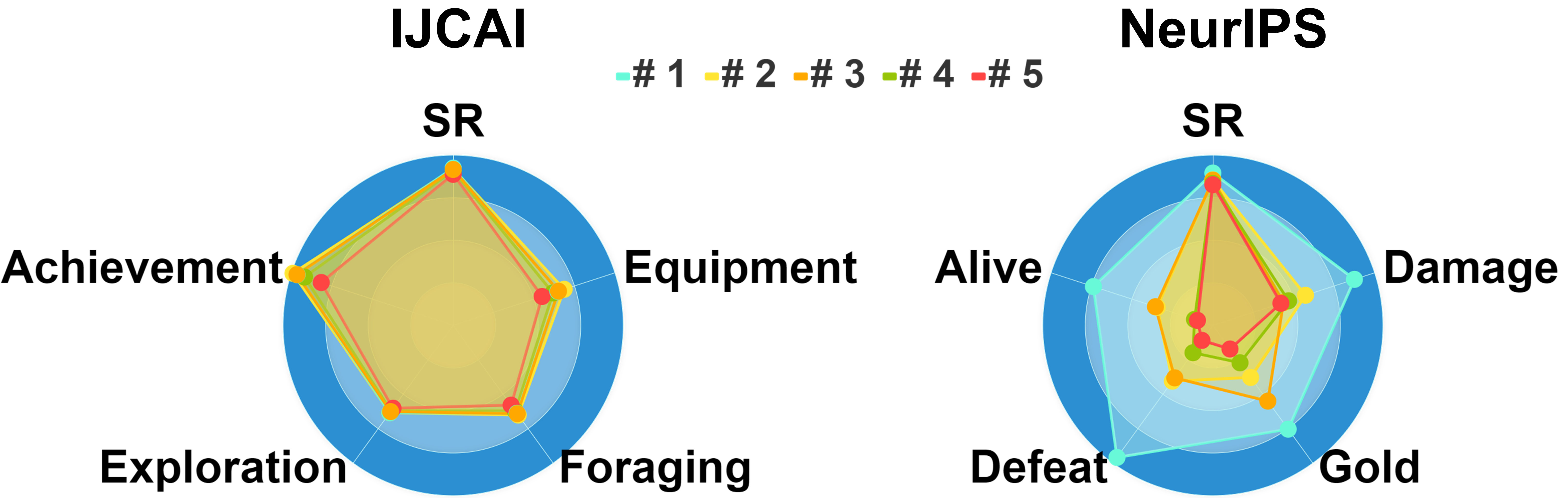}
    \caption{An overview of top 5 submissions' capabilities in the past two competitions.}
    \label{radar}
\end{figure}

It is worth noting that the change in the scoring mechanism in this competition resulted in a better representation of the strengths and weaknesses of participants' policies. As shown in Fig. \ref{radar}, in the previous competition, the top 5 participants' scores were very close, and it was difficult to distinguish their strategies based on the achievement indicators. In contrast, in this competition, although the top 5 participants' results were similar, we were able to distinguish obvious differences among their policies based on evaluation indicators.

To summarize, the game mechanics of this competition were more interesting than those of the previous competition, and the flexible scoring system showcased the potential of the Neural MMO environment for MARL research.

\begin{figure}[!ht]
    \centering
    \includegraphics[width=5.5in]{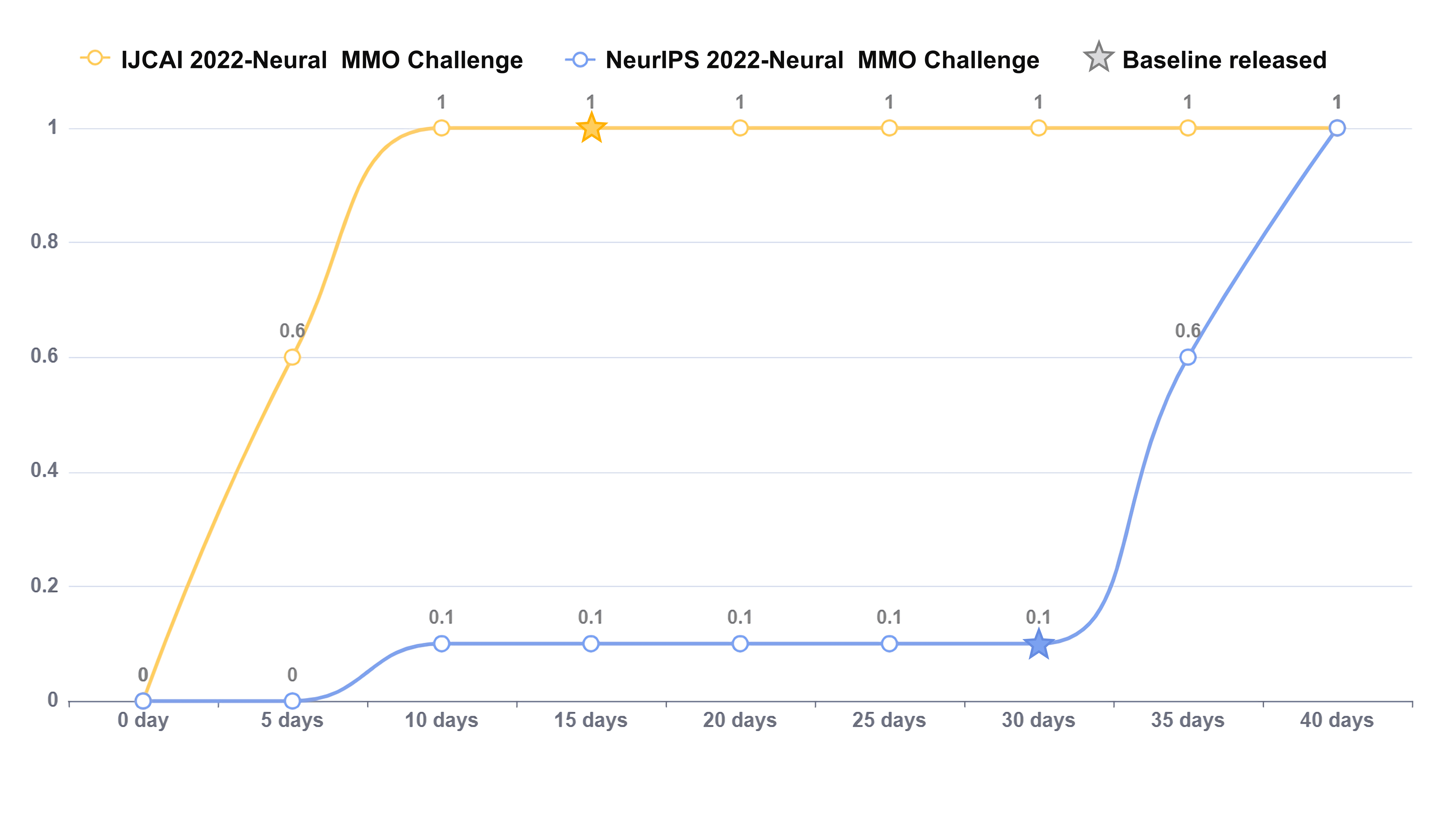}
    \caption{Submission performance over time in PvE Stage 1 of the past two competitions.}
    \label{baseline}
\end{figure}

\subsubsection{Tools and Service}

Figure \ref{baseline} shows the submission performance curve of the NeurIPS-2022 Neural MMO PvE stage 1. Unlike the previous competition, where participants achieved ideal results without a baseline, this time, the number of submissions was infrequent, and none of them achieved good results before we provided the baseline. This indicates that as competition difficulty increases, participants require additional guidance. Therefore, we launched our RL baseline, which resulted in an increase in the frequency of submissions and breakthroughs in submissions by participants.

Apart from providing a baseline, we also offered a starter kit, Q\&A, and live broadcasts to help participants better understand the competition's progress and details, thereby improving their policies.

In summary, our tools and services enable participants to concentrate on implementing their ideas and refining their policies rather than spending time building from scratch.

\section{Conclusion}
We organized the NeurIPS 2022 Neural MMO challenge, which attracted over 500 participants from 35 countries and regions. The competition featured two tracks, PvE and PvP, and we observed that the same policy may perform differently on both tracks. In PvE, the competition is designed with a difficulty ladder. Most participants achieve a top1 ratio of 1 in the first phase, but only three players can achieve a top1 ratio of 1 in the second phase, and the score difference between participants is significant. This highlights the potential of this environment as a benchmark for algorithm robustness and generalization.

The environment introduces various mechanics, such as trading, skill differentiation, and many-agent cooperation, making it more interesting and suitable for MARL research. Analyzing the item distribution in the environment, we found some human-like trading behaviors, which surprised us.

Before the baseline was provided to participants, the number of submissions was minimal, and the majority of participants received unsatisfactory scores. This phenomenon reflects the increased difficulty of the environment. However, after the baseline was released, we observed an uptick in the number and quality of submissions. This highlights the importance of tools and services surrounding the environment for RL research, rather than relying solely on algorithmic innovation. We believe that with the advent of generic and flexible tools, there is room for substantial improvement to policy robustness and generalization on Neural MMO.

\bibliography{nips}

\appendix

\section{Environment Setting for NeurIPS Neural MMO Challenge}\label{apd:first}
\subsection{Environment Wrapper}

To make the environment more extensible and participant-friendly, we have incorporated some new functions in addition to the original environment. These changes include:

\begin{enumerate}
    \item Random assignment of team numbers and spawns, which allows teams start at different initial positions on the map after multiple evaluations. This is helpful for addressing asymmetry in evaluation of participants' algorithms.
    \item Aggregation of observations of single agents at the team level, allowing each agent to have the information of other teammates. Our approach is different from OpenAI Five and AlphaStar as it does not strictly conform to the MARL definition within an individual team.
\end{enumerate}

\subsection{Tracks}

As shown in Fig. \ref{track}, the competition is divided into two tracks: PvE and PvP. In the PvE track, participants' policies are evaluated against 15 built-in policies (i.e. copies of policies that we trained), while in the PvP track, they are evaluated against policies submitted by 15 other participants. From a competitor's perspective, the PvE track is designed to help players quickly familiarize themselves with the environment against a fixed set of opponents, while the PvP track is intended to increase the competitiveness of the game. The structure of the different tracks is as follows:

\begin{figure}[!ht]
    \centering
    \includegraphics[width=\linewidth]{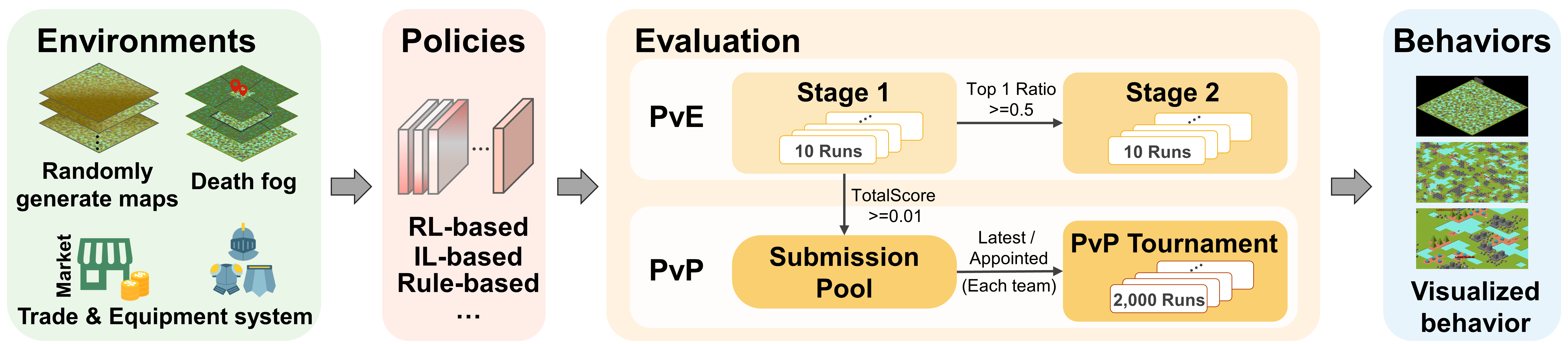}
    \caption{An overview of the track structure and evaluation process of the competition.}
    \label{track}
\end{figure}

\subsubsection{PvE Stage 1 vs Scripted AI}
In this stage, participants' policy will be thrown into a game with 15 scripted built-in AI teams and evaluated for 10 rounds. The result of each submission will be presented as a Top 1 Ratio (see section \ref{Metrics}). Each built-in AI team (Mixture Team and Combat Team) in stage 1 consists of 8 agents with 8 respective skills. These policies are open-sourced so that an evaluation environment is accessible during training. Each member of the mixture team specializes in a specific skill. The combat team is hostile and will attack every agent close to them. The aim of this track is to help new participants get familiar with the challenge.

\subsubsection{PvE Stage 2 vs RL-based AI}

To qualify for Stage 2 in the PvE track, a participant must achieve a 0.4 Top 1 Ratio in Stage 1. In Stage 2, the rules are the same as in Stage 1, except that the built-in AIs are designed by the organizers and trained using a deep reinforcement learning framework developed by Parametrix.AI. This stage is intended to be more challenging than Stage 1, as the built-in AIs are much stronger, and defeating them is a significant achievement. There are three types of built-in AI teams in Stage 2:

\begin{enumerate}
    \item Reckless: These bots are skilled in combat and also in using, selling, and buying items.
    \item Ruthless: An evolution of the Reckless policy, Ruthless bots are both hostile and cooperative. They will fight enemies independently but also recognize the importance of team cooperation.
    \item Coward: These bots are risk-adverse and avoid combat with high-level NPCs. They move quickly to the edges of the map to ensure a survival advantage and snipe incoming opponents for points.
 \end{enumerate}

\subsubsection{PvP}
In this stage, each participant's policy is pitted against the policies sampled from the pool of qualifying submissions. Given the number of submissions from different teams employing different approaches, this procedure is an effective assessment of robustness and generalization. To ensure an accurate assessment, each participant will play at least 1000 matches.

\subsection{Metrics}\label{Metrics}
\subsubsection{Score}
The participants' policy score is determined by the number of other agents (excluding NPCs) they defeat (i.e., inflict the final blow) and how long at least one team member remains alive. Each agent defeated earns 0.5 points for the participants' policy, and the final defeat score is the sum of the per-agent defeat scores across the team. The team's survival time is determined by when the last agent dies. The survival scores in each tournament range from 10 to 0. Specifically, the survival scores in each tournament from high to low are 10, 6, 5, 4, 3, 2, 1, 1, 0, 0, 0, 0, 0, 0, 0, 0, and will be assigned to the team surviving the longest, the second longest, the third longest, and so on respectively. In the case of tied survival times, the team with more survivors ranks higher. If the number of survivors is equal, the average level of the agents is taken into account. If all factors are still equal, the tied teams will split the ranking scores. For example, if a policy lands the last hit on 5 agents and survives the longest, they receive 5*0.5+10 = 12.5 points.

\subsubsection{Top 1 Ratio in PvE}
The PvE track's fixed opponents offer a consistent performance metric for both scripted and reinforcement learned submissions. A Top 1 Ratio of 1.0 over 10 games (i.e. 10 wins in a row) indicates that the submission is significantly better than the built-in AI. This makes Top 1 Ratio a reasonable first screen of the performance of competitors.

\subsubsection{TrueSkill in PvP}
Top 1 Ratio is a poor indicator of performance in PvP evaluations against unknown opponents. We instead use a TrueSkill scoring mechanism that takes into account the ranking of all policies in a match, not just the winner. In other words, even if a strategy loses first place in a competition, its ranking under TrueSkill can still improve.

\subsection{Resources}

\subsubsection{Starter Kit}
This project includes environment installation instructions, submission demos, and FAQs for common problems to ensure that participants have a positive experience. Everything necessary for participants to submit their work is included.

\subsubsection{Baseline}
To accommodate the limited device resources of most participants, we have developed a baseline using torchbeast that is available to all participants. This baseline enables participants to achieve a 0.5 Top 1 ratio on a PC with a GPU within one day.

\subsubsection{Environment Documentation}
The documentation provided by the Neural MMO website is intended for a broader audience of researchers using the base Neural MMO platform for a variety of research agendas. This makes it difficult for competitors to get the information they need in a short time. To help participants get started quickly, we have published a competition-specific document containing important information about the environment.

\subsubsection{Web Viewer}
\begin{figure}[!t]
    \centering
    \includegraphics[width=\linewidth]{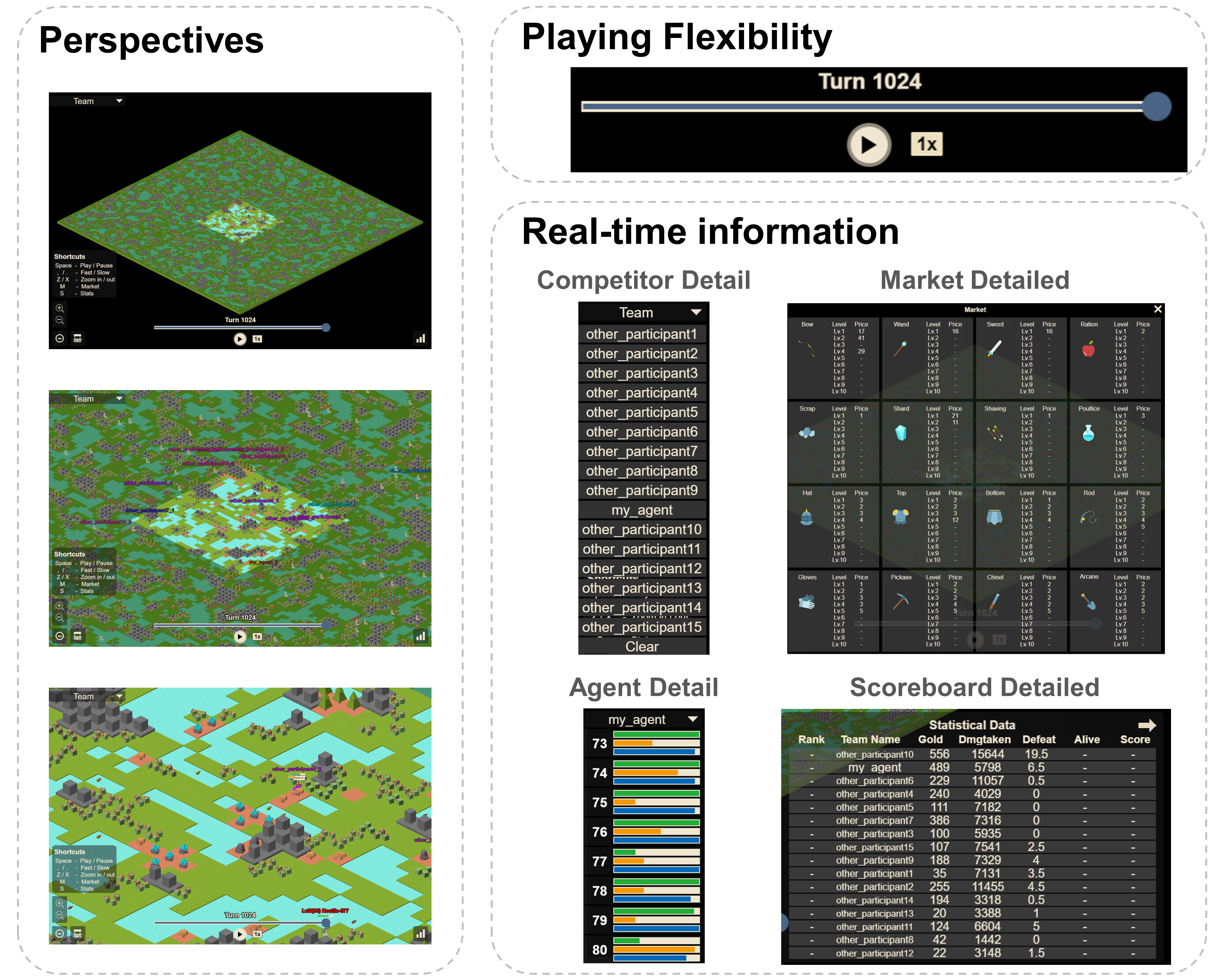}
    \caption{An overview of web viewer. Left: various zoom levels. Top Right: draggable timeline. Bottom Right: Detailed statistics about the agent and market states.}
    \label{webviewer}
\end{figure}

We have created a browser-based video replay tool that allows participants to view their policy performance in terms of movement, collection, fighting, and trading, enabling them to make better policy adjustments. The web viewer includes different perspectives, controllable playback speed, and detailed information for display, as shown in Fig.~\ref{webviewer}.

 \subsubsection{Evaluation System}
 We have designed a distributed evaluation system based on k8s to handle large-scale submissions and fully evaluate the performance of each player's policy. During the PvE phase, each player's strategy will compete against built-in AI on a randomly generated map. In the PvP phase, all players in PvE stage 2 will compete against each other. For the PvP track, we provide both daily and weekly evaluations. Daily evaluations run only 100 matches against other participants. This functions as a more frequent feedback signal to help participants improve their policies. We also run a more extensive 1000 match evaluation once per week. As the performance of a policy is always relative opponents, this system helps to provide an accurate evaluation of player strategies.

\section{Policy Analysis}\label{apd:second}
\subsubsection{Different Types Algorithms}

\begin{figure}[!b]
    \centering
    \includegraphics[width=6in]{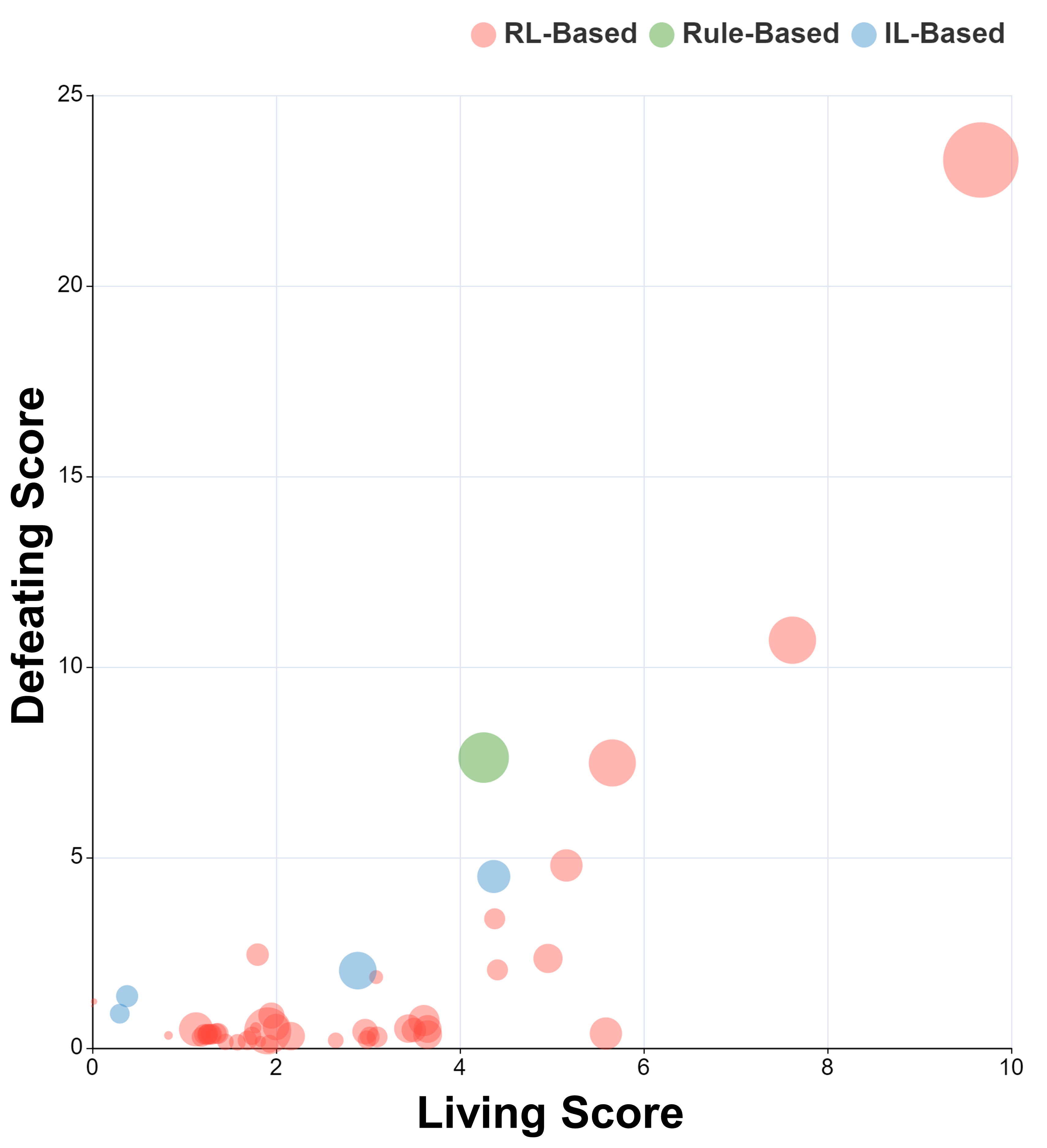}
    \caption{Performance of different type algorithms in the NeurIPS-2022 Neural MMO challenge PvP stage. Every point in the chart denotes a participant. The size of the point represents the total number of gold obtained by that participant.}
    \label{bubble}
\end{figure}

We collected over 1600 policies and sorted participants into three categories based on the algorithms they used: rule-based, RL-based, and IL-based. Fig. \ref{bubble} shows a comparison between these three types of algorithms. As depicted, most participants chose the RL algorithm for the challenge, and some of them achieved a top 16 ranking. However, a few participants submitted scripted policies, and one of the IL submissions achieved 5th place.

Overall, RL took participants longer to get up and running but delivered higher performance by the end of the competition. The theoretical upper bound of IL-based learning is close to that of RL but is limited by the quality of the data (which, in this case, is generated by an RL policy because human data is unavailable). Rule-based methods allow participants to control the actions of agents directly. We noticed that scripted approaches were quick to develop and effective at the start of the competition, but their performance fell off compared to RL by the end, and they were overly adapted to the opponents of the first two PvE stages.

\subsubsection{Different Tracks}
\begin{figure}[!ht]
    \centering
    \includegraphics[width=6in]{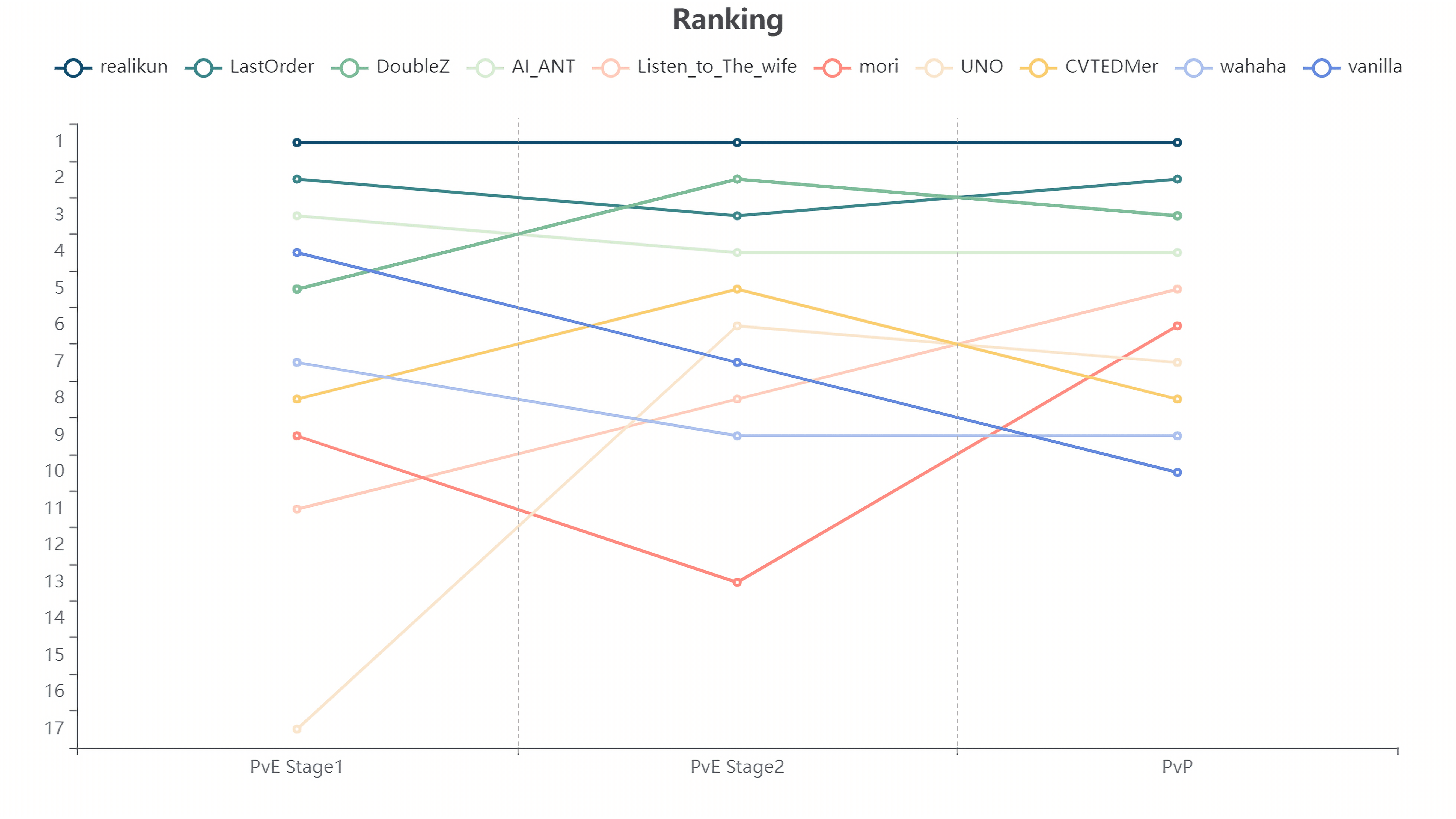}
    \caption{The ranking of participants in different tracks.}
    \label{track2}
\end{figure}

The main challenges of this competition are robustness and generalization to new opponents in a complex environment. In the NeurIPS-2022 Neural MMO challenge, we set up two different tracks that can fully test the robustness and generalization of participants' algorithms. As shown in Figure \ref{track2}, the top 3 participants in PvE stage 1 all demonstrate good robustness and generalization. Realikun consistently wins in all tracks, while the rankings of the other two players remain steady. However, for most other participants, their rankings change continually. For example, the Vanilla algorithm is weak on generalization, performing well in PvE stage 1 but poorly in PvE stage 2 and PvP stage. These observations demonstrate that this environment can facilitate the evaluation of policy robustness and generalization by introducing different opponents. They also illustrate that when a policy is robust enough, it is able to outperform all other policies in the environment.

\subsubsection{Different Combat Preferences}
After analyzing Fig. \ref{radar}, we observed that even strategies with very close total scores behaved differently. We focus on the analysis of the strategies used by the top three players. As depicted in the figure, it is evident that realikun aimed to obtain higher points by defeating more participants. Typically, most participants believe that the maximum survival score is 10 points and that achieving this score will create a significant lead over their opponents. As a result, most participants tend to be relatively conservative and avoid fighting aggressively to attain this score. However, if most participants are conservative, they may be at a disadvantage when fighting becomes inevitable later in the game. This allows a more aggressive policy to become substantially stronger than most participants, ensuring that it can defeat all other agents and survive to the end. Evidently, realikun is based on this idea and successfully won first place in all tracks of the competition. In contrast to realikun, doublez avoids unnecessary fighting (rarely fighting powerful NPCs in the environment), while passerby's agents are more aggressive. Passerby has more combat points than doublez, but it takes more damage and has lower survival rate, as shown in the figure.

% \begin{figure}[!b]
%     \centering
%     \includegraphics[width=5.5in]{./parallel.png}
%     \caption{Indicators for a participants' performance. The green line represents ikun, who win the first place. while the pink lines denote passerby and doubles, respectively}
%     \label{parallel}
% \end{figure}

\begin{figure}[t]
    \centering
    
    \subfigure[Distribution of items]{
        \centering
        \includegraphics[width=1.8in]{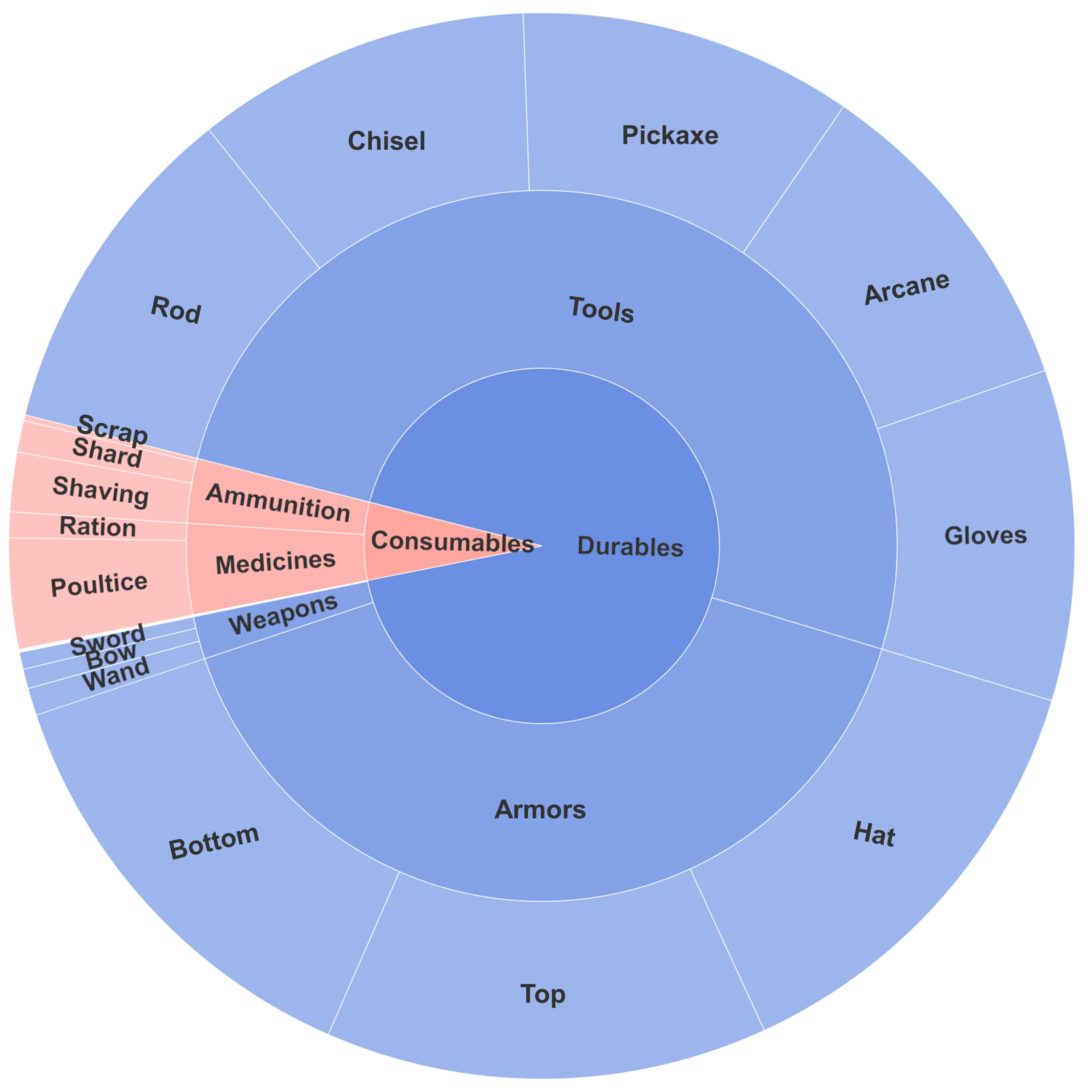}
    }
    \subfigure[Quantity of items]{
        \centering
        \includegraphics[width=1.8in]{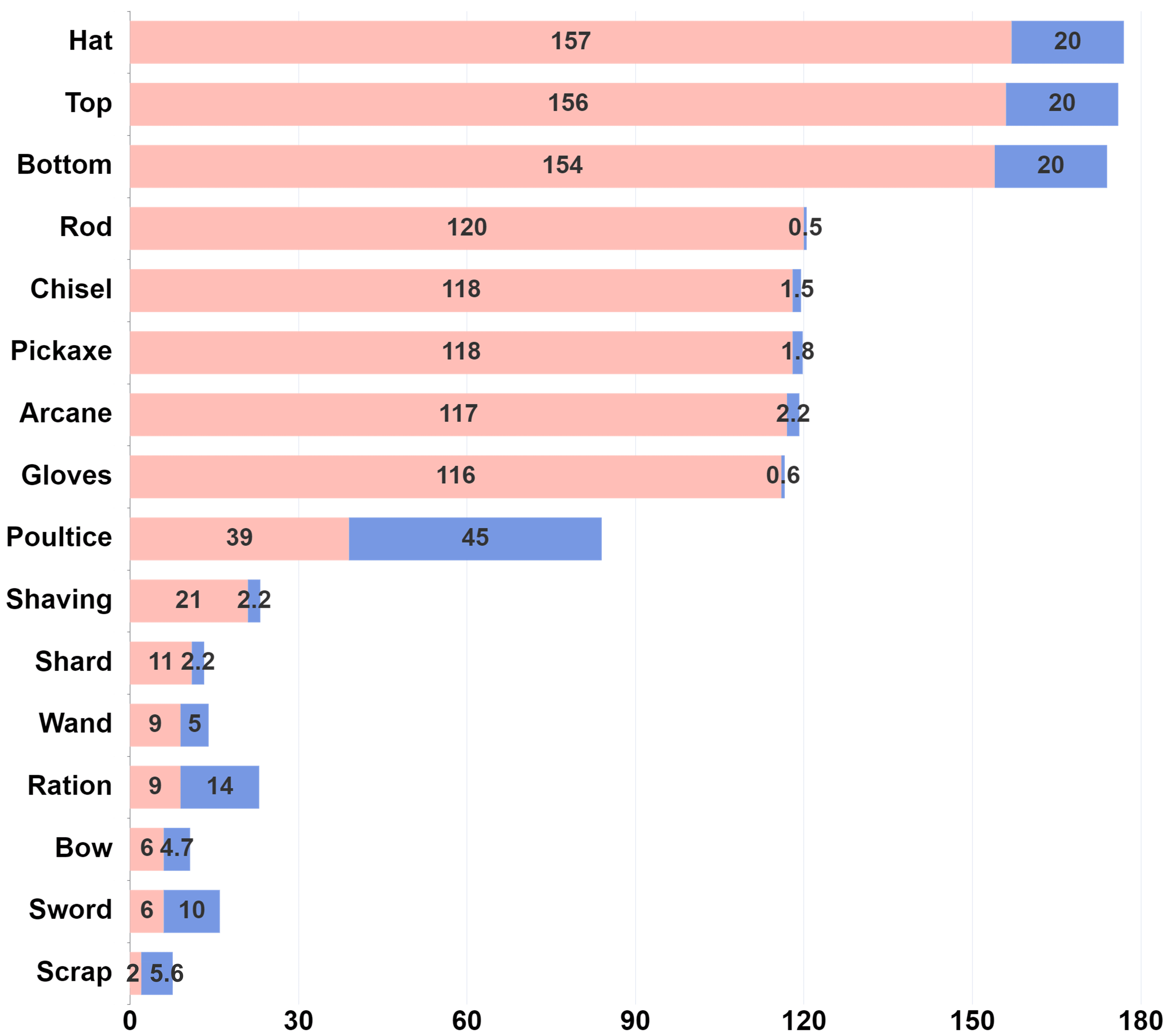}
    }
    \subfigure[Buy/sell ratio]{
        \centering
        \includegraphics[width=1.8in]{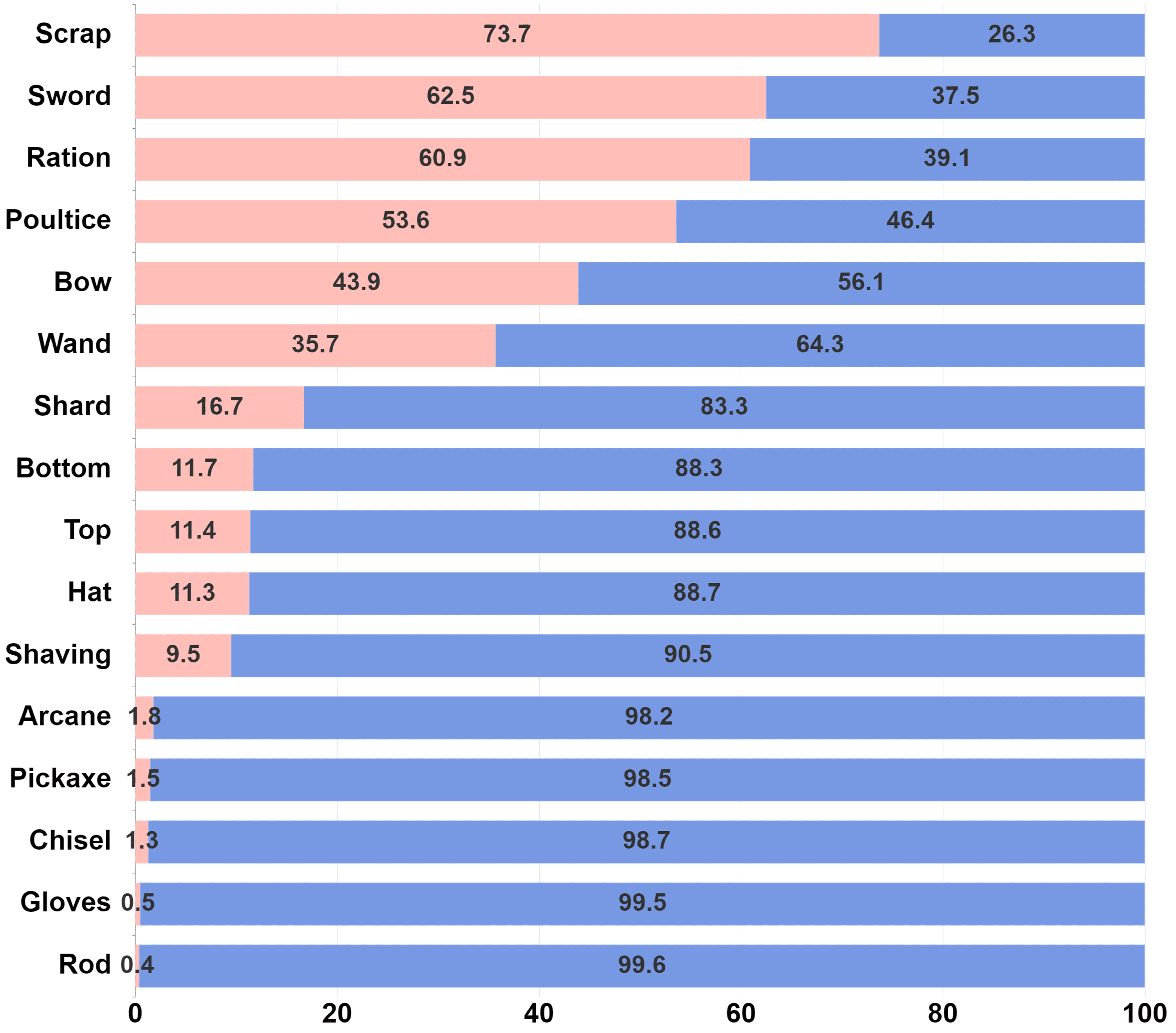}
    }

    \caption{Details of the distribution of items in the NMMO environment.}
    \label{trade}
\end{figure}

\begin{figure}[h]
    \centering
    \includegraphics[width=\linewidth]{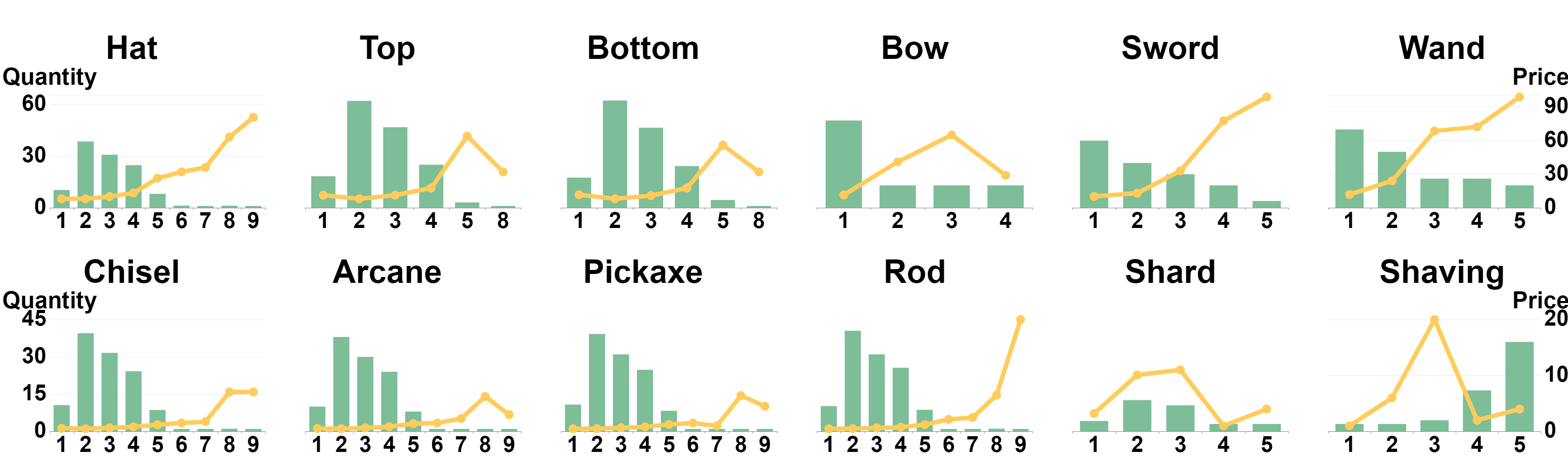}
    \caption{Quantity available and price of each item at various levels. Acquiring higher-level items is generally more difficult.}
    \label{grid}
\end{figure}

\subsubsection{Trade Overview}
Figure \ref{trade} (a) shows the distribution of items in the environment, which has numerous tools and few consumables. Additionally, Figure \ref{trade} (b) depicts the amount of items, revealing that the total number of items of the same type is comparable, indicating the environment's setting is reasonable. Based on this, we can deduce that different items hold different importance for agents, as shown in Fig. \ref{trade} (c). Agents prioritize owning weapons and restoration potions rather than selling them to the global market. However, since agents' inventory is constrained, they need to trade their some supplies in the global market to make space. As seen in Fig. \ref{grid}, we observe some interesting trading strategies: (1) For universally preferred items (such as weapons), agents' offers usually increase with the level (difficulty to obtain) of the weapon. (2) Similarly, the value of such as tools also increases with their level, but if agents realize the expected price is too high to sell top-level supplies, they may reduce the price for trading.

\end{document}